\documentclass[lettersize,conference]{IEEEtran}
\IEEEoverridecommandlockouts

\usepackage{cite}
\usepackage{amsmath,amssymb,amsfonts}
\usepackage{algorithmic}
\usepackage{graphicx}
\usepackage{textcomp}
\usepackage{xcolor}
\usepackage{url}
\usepackage{booktabs}
\usepackage{multirow}
\usepackage{pifont}
\usepackage{tikz}
\usepackage[caption=false,font=footnotesize]{subfig}

\def\BibTeX{{\rm B\kern-.05em{\sc i\kern-.025em b}\kern-.08em
    T\kern-.1667em\lower.7ex\hbox{E}\kern-.125emX}}

\begin{document}

\title{SynCalib: Breaking Dataset Shortcuts in LiDAR-Camera Calibration via Extrinsic-Controllable Synthesis}

\author{Huihua Zhu\textsuperscript{1}, Shuyang Zhang\textsuperscript{1,$\dagger$},
  Dongqing Chen\textsuperscript{1}, Zhuo Yu\textsuperscript{1},
  Jianhao Jiao\textsuperscript{2}, \\
  Yilong Zhu\textsuperscript{3}, Siyu Teng\textsuperscript{4},
  Xiaopin Zhong\textsuperscript{1,$\dagger$}, and Zongze Wu\textsuperscript{1}
  \thanks{\textsuperscript{1} Huihua Zhu, Shuyang Zhang, Dongqing Chen, Zhuo Yu,
    Xiaopin Zhong, and Zongze Wu are with the College of Mechatronics and Control
    Engineering, Shenzhen University, Shenzhen, Guangdong, 518060, China.
    ~(Corresponding author: Shuyang Zhang, Xiaopin Zhong)}
  \thanks{\textsuperscript{2} Jianhao Jiao is with the Robot Perception and Learning
    Lab, Intelligent Robotics, Department of Computer Science, University College
    London, Gower Street, WC1E 6BT, London, UK.}
  \thanks{\textsuperscript{3} Yilong Zhu is with the Department of Electronic and
    Computer Engineering, Hong Kong University of Science and Technology, Hong
    Kong.}
  \thanks{\textsuperscript{4} Siyu Teng is with the College of Civil and Transportation 
    Engineering, Shenzhen University, Shenzhen, Guangdong, 518060, China.}
}

\maketitle

\begin{abstract}
  Online LiDAR-camera calibration remains a fundamental yet challenging task in the 
  field of robotics. 
  Recently, deep learning-based data-driven approaches have emerged, offering high-precision, 
  targetless calibration capabilities. 
  However, existing methods often suffer from a critical \textit{dataset shortcut} phenomenon 
  due to the lack of extrinsic diversity in their training data. 
  Networks tend to overfit to fixed nominal extrinsic configurations, 
  leading to severe performance degradation and poor generalization in real-world 
  deployments. 
  To address this limitation, we propose a highly generalizable data generation 
  framework tailored for training LiDAR-camera calibration networks. 
  Our framework leverages existing omnidirectional image datasets annotated with depth 
  ground truth to synthesize multi-modal calibration data under rigorous extrinsic 
  constraints. 
  Extensive experiments demonstrate that the data generated by our pipeline significantly 
  enhances the generalization and robustness of typical learning-based calibration 
  networks. 
  Furthermore, we conduct comprehensive real-world evaluations across various camera-LiDAR 
  configurations, validating the zero-shot transfer capability of models trained 
  exclusively on our synthetic data. 
  To facilitate future research and community use, the code for data generation, the 
  pre-trained models, and online benchmark datasets will be made publicly available.
\end{abstract}

\section{Introduction}
Multi-modal systems have become indispensable for modern robotics, with the combination
of cameras and LiDARs serving as the predominant
configuration~\cite{lang2026ijrr, xue2026tase, li2026ral}.
Accurately estimating the extrinsic parameters between these sensors to achieve
precise spatial alignment is a critical prerequisite for downstream tasks such as
perception~\cite{liu2022icra}, navigation~\cite{zheng2024tro}.

During manufacturing, robots typically undergo precise extrinsic calibration in
dedicated facilities using specialized fiducial markers (e.g., checkerboards).
While these target-based methods achieve exceptional accuracy by leveraging known
geometric priors, sensors deployed in the field inevitably experience extrinsic
drift due to mechanical vibrations and thermal stresses in harsh operational
environments~\cite{koide2023icra}.
Returning robots to the factory for re-calibration incurs prohibitive maintenance
costs and system downtime.
Consequently, targetless online calibration, which estimates extrinsics directly
from natural scenes without specialized facilities, has emerged as a practical
alternative.
Despite being marginally less accurate than target-based approaches, its operational
convenience and scalability make it highly desirable for mass-produced robotic systems.

\begin{figure}
  \centering
  \includegraphics[width=0.95\columnwidth]{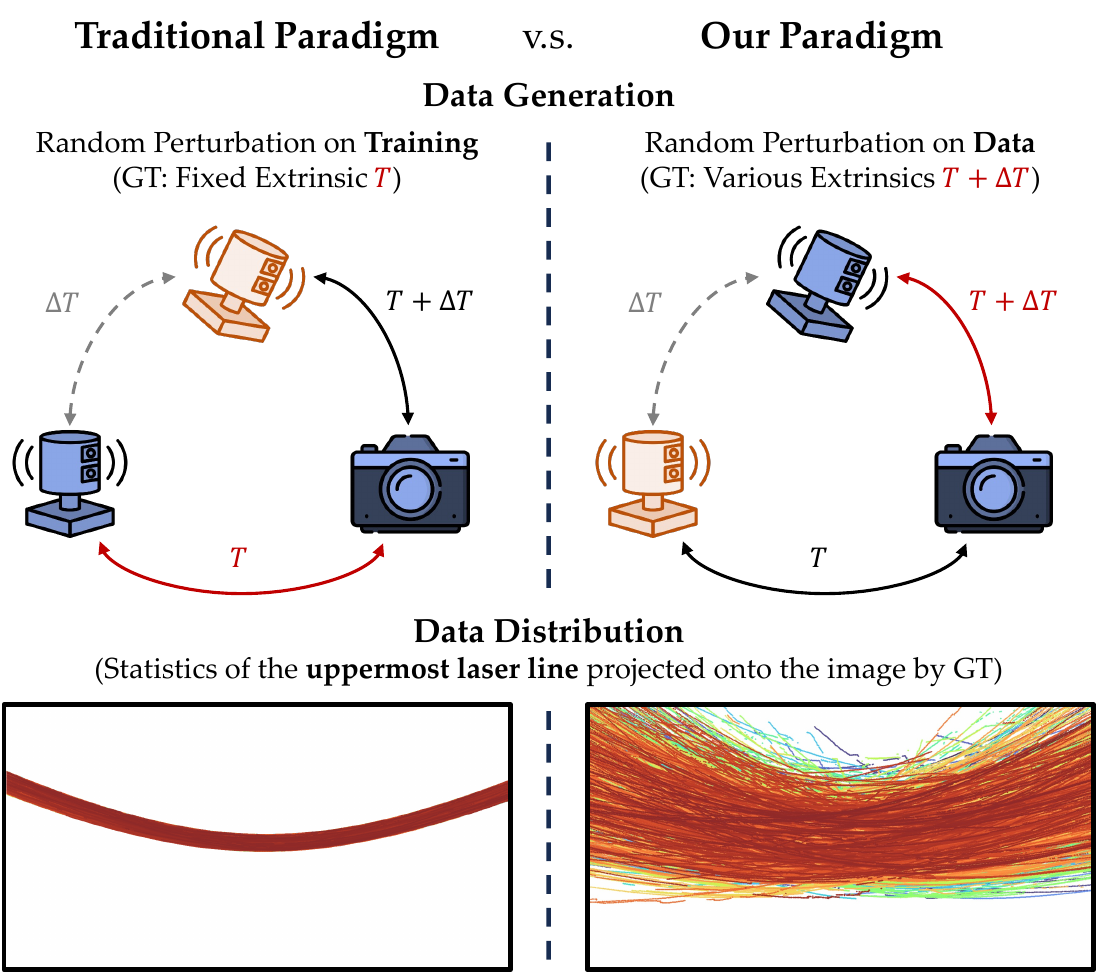}
  \caption{Comparison between traditional and our data paradigm.
    \textit{Traditional methods} use public datasets, which provide training data with
    a fixed LiDAR-camera extrinsic, while \textit{ours} generates pairs with various
    configurations. The data visualization shows that our training data has richer patterns
    (projections after calibration), while the traditional one introduces harmful inductive bias.}
  \label{fig_cover}
\end{figure}

Deep learning-based calibration methods have emerged as the dominant paradigm
in recent years, for its adaptability to various scenarios and excellent performance.
However, these methods suffer from a critical flaw of severe dataset bias~\cite{huang2025acmmm,cattaneo2025tro,huang2026tro}.
Recent models are typically trained on several public datasets
(e.g., KITTI~\cite{geiger2013ijrr} and nuScenes~\cite{caesar2020cvpr}) with highly
restricted sensor setups.
This scarcity of diverse extrinsic configurations leads to severe representational
homogeneity in the training data (Fig.~\ref{fig_cover}).
Consequently, instead of learning meaningful cross-modal associations, network
resorts to \textit{shortcut learning}: It merely memorizes the spatial distribution
priors of LiDAR projections to minimize training loss.
Ultimately, this causes existing algorithms to overfit to the specific LiDAR distribution
and fail to provide reliable extrinsics in practice.

To tackle this challenge, our core insight is to systematically diversify the extrinsic
ground truths in the training data, thereby eliminating the conditions that foster
shortcut learning.
To this end, we propose SynCalib, a novel offline data synthesis framework
designed for arbitrary LiDAR-camera configurations.
By leveraging existing simulated environments equipped with omnidirectional cameras,
SynCalib synthesizes large-scale LiDAR-camera data pairs with fully controllable
extrinsic parameters.
This guarantees the richness of extrinsic configurations during network training,
fundamentally enhancing the model's zero-shot generalization capabilities.
Our main contributions are summarized as follows:
\begin{itemize}
  \item A calibration data generation framework that enables model
        training prior to physical sensor deployment.
        Models trained with our synthesized data can be deployed in a zero-shot manner,
        and effectively circumvent shortcut learning.
  \item An extrinsics-controllable synthesis pipeline leveraging a simulated
        omnidirectional visual system with dense depth ground truth.
        This pipeline features fully controllable extrinsic sampling to ensure high-fidelity
        data synthesis.
  \item Comprehensive experiments demonstrating that our system successfully overcomes
        the performance degradation caused by dataset bias in traditional training paradigms.
\end{itemize}

\section{Related Works}
LiDAR-camera calibration methods are generally categorized into \textit{target-based}
and \textit{target-free} approaches.

\subsection{Target-based LiDAR-camera Calibration}
Target-based methods rely on specialized calibration objects with distinct semantic
or geometric structures.
Alignment is achieved by extracting cross-modal features and applying geometric solvers
like Perspective-n-Point~(PnP).
Typically, planar boards are used, where precise manufacturing guarantees the spatial
relationship between visual and geometric cues~\cite{fang2021iros, jiao2023tmech, zheng2026rap}.
Xie~\textit{et al.}~\cite{xie2022ral} designed a 3D depth-textured target for solid-state
LiDARs to ensure salient feature extraction.
Chen~\textit{et al.}~\cite{chen2023ral} aligned image color with LiDAR reflectivity using
a textured planar scene, optimizing extrinsics via plane-constrained bundle adjustment.

Despite their exceptional precision, these methods require custom targets, controlled
environments, and empirical expertise for data collection.
Consequently, they are labor-intensive and costly, often necessitating factory re-calibration
or professional operators.

\subsection{Target-free LiDAR-camera Calibration}
Target-free approaches leverage natural environmental content for on-the-fly calibration.
Traditional methods establish cross-modal associations via mutual information~(MI)
maximization~\cite{pandey2015jfr, stutts2024icra} or edge alignment~\cite{hu2023icra, ye2024iros},
followed by geometric optimization.
Recent works incorporate semantic segmentation~\cite{huang2024tiv},
deep feature matching~\cite{koide2023icra, wang2025taes}, and neural rendering~\cite{jung2026ral}
for more robust associations.
Additionally, ego-motion and hand-eye calibration frameworks~\cite{ou2023sj, petek2024ral}
enable extrinsic estimation even with minimal field-of-view~(FoV) overlap.

Recently, end-to-end neural networks~\cite{zhu2023icra, duan2023ral, ou2026ral}
have become mainstream, directly regressing relative poses by implicitly learning
alignment cues from large-scale datasets.
However, regression-based pose learning is highly susceptible to scene
overfitting~\cite{sattler2019cvpr} and shortcut learning~\cite{huang2025acmmm},
often exploiting dataset biases rather than learning genuine cross-modal geometric
alignment.

\subsection{Practical Solutions for Dataset Shortcuts}
To address shortcut learning, recent strategies aim to eliminate dataset biases.
CMRNext~\cite{cattaneo2025tro} mitigates this by replacing end-to-end regression
with a hybrid framework, using RAFT-based dense matching for feature correspondences
followed by a PnP solver.
DST-Calib~\cite{huang2026tro} identified this issue and proposed
transforming the problem into a geometry-to-geometry registration task by utilizing
a monocular vision model (e.g., MoGe-2~\cite{wang2026neurips}) to convert RGB images
into depth maps.
Furthermore, they implemented data augmentation on both the image and LiDAR branches,
leveraging cross-frame matching to enhance extrinsic generalization.

Unlike these algorithmic interventions, our work tackles shortcut learning from a
purely \textit{data-centric} perspective.
We utilize simulated omnidirectional images to synthesize large-scale LiDAR-camera
training pairs with highly diverse extrinsic constraints.
Notably, our primary focus lies exclusively in the generation of high-quality calibration
data, without requiring modifications to the underlying architectures of existing
deep learning networks.
By training mainstream calibration models directly on our synthesized data, we
empirically demonstrate the elimination of dataset shortcuts and a significant
improvement in zero-shot generalization.

\begin{figure}
  \centering
  \subfloat[
  Comparison of model inference pipelines between the traditional paradigm and 
  our proposed system. 
  We introduce extrinsic perturbations during the data generation phase and initialize 
  the network input using fixed factory extrinsics, which closely reflects real-world 
  deployment scenarios.
  ]{
    \centering
    \includegraphics[width=0.97\columnwidth]{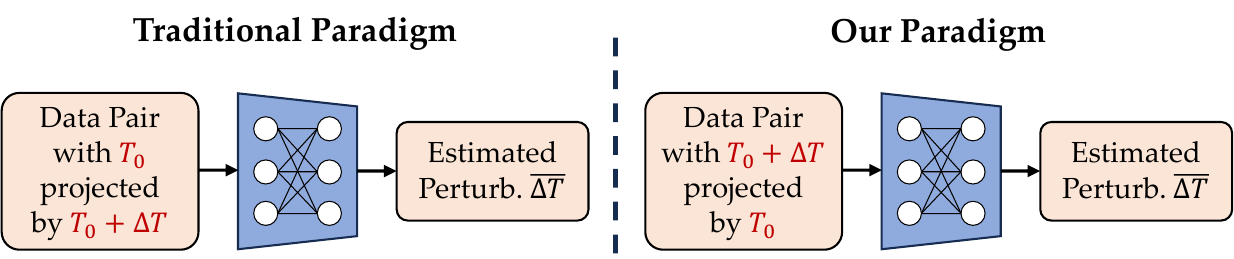}
    \label{fig_problem_formulation}} \\
  \vspace{-0.2cm}
  \subfloat[The system overview of our data generation process.]{
    \centering
    \includegraphics[width=0.95\columnwidth]{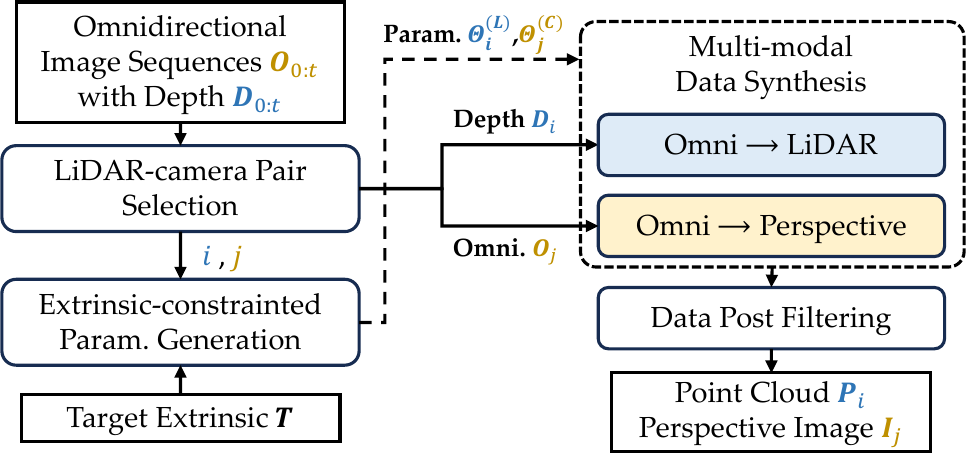}
    \label{fig_overview}}
  \caption{Our training paradigm and data generation pipeline.}
  \label{fig_problem_overview}
\end{figure}

\section{Methodology}
\subsection{Problem Formulation and System Overview}
LiDAR-camera calibration networks typically estimate extrinsic residuals
$\Delta \mathbf{T} \in \mathrm{SE}(3)$ from input image $\mathbf{I} \in \mathbb{R}^{W \times H \times 3}$
and point cloud $\mathbf{P} \in \mathbb{R}^{N \times 3}$.
During the training stage, they project point cloud $\mathbf{P}$ onto the image $\mathbf{I}$
as depth map $\mathbf{D} \in \mathbb{R}^{W \times H}$ using ground-truth extrinsics
$\mathbf{T}_{0}$ and apply perturbations $\Delta \mathbf{T}$ at the input stage.
However, fixed hardware configurations in public datasets severely limit extrinsic
diversity, and this conventional projection creates learning shortcuts via highly
consistent spatial features.

To address this, we introduce extrinsic perturbations $\Delta \mathbf{T}$ during
the data generation phase rather than the input stage (Fig.~\ref{fig_problem_formulation}).
We first generate image $\mathbf{I}$ and point cloud $\mathbf{P}$ pairs with the
ground truth extrinsics $\mathbf{T} + \Delta \mathbf{T}$.
For the network input, we project the point cloud using the nominal transformation
$\mathbf{T}$, which is defined as the factory extrinsic in practice.
This novel formulation naturally enriches extrinsic diversity and effectively
eliminates learning shortcuts.

The overview of our system is illustrated Fig.~\ref{fig_overview}, which generates
LiDAR-camera training data using simulated datasets with ground-truth depth.
The methodology unfolds as follows:
Section~\ref{subsect_data_synthesis} details multi-modal data generation;
Section~\ref{subsect_par_generation} formulates extrinsic constraints between modalities;
and Section~\ref{subsect_post_process} describes surround-view data selection
and post-filtering to remove low-quality samples.

\subsection{Multi-modal Data Synthesis}
\label{subsect_data_synthesis}
We synthesize perspective images and LiDAR point clouds under arbitrary viewpoints
and sensor configurations from omnidirectional images.
We utilize the TartanGround~\cite{patel2025iros} dataset, which contains full $360^{\circ}$ scene 
images with ground truth depth.
These images are stored by a CubeMap format, consisting of six mutually orthogonal
views, each with a FoV of $90^{\circ}$.
Without loss of generality, we convert these CubeMaps into equirectangular projection~(ERP)
images to facilitate subsequent operations.
The ERP format maps the spherical surface onto a projection image with a $2$:$1$ aspect ratio,
where each pixel strictly corresponds to a spatial angular $\{\theta,\phi\}$.
Detailed operations regarding ERP images can be found in \cite{huang2022360vo}.

\subsubsection{Omni2LiDAR}
Mainstream LiDARs are broadly categorized into repetitive scanning~(including solid-state LiDARs)
and non-repetitive scanning types.
By defining a temporal list of rays, LiDAR sensors measure the corresponding depth
of each ray based on the Time-of-Flight~(ToF) principle, which is subsequently
converted into a point in Euclidean space.

We construct a LiDAR pattern bank by importing the parameter specifications of
mainstream LiDAR sensors.
Repetitive scanning LiDARs show constant ray angles.
Specifically, for non-repetitive scanning LiDARs (e.g., the Livox Mid-360)
characterized by complex spatial angular sequences, we extract a continuous $4$-second
directional angle sequence from official simulation examples.
By randomly sampling a start time $t_{start}$ from the sequence, we obtain non-repetitive
scanning patterns with a fixed duration of $d$ seconds.
Furthermore, to account for the physical limitations of LiDAR sensors, such as
maximum depth estimation ranges, we simply apply depth range thresholds
$\{d_{\text{min}},d_{\text{max}}\}$ to filter out-of-range points.

\begin{figure}
  \centering
  \includegraphics[width=\columnwidth]{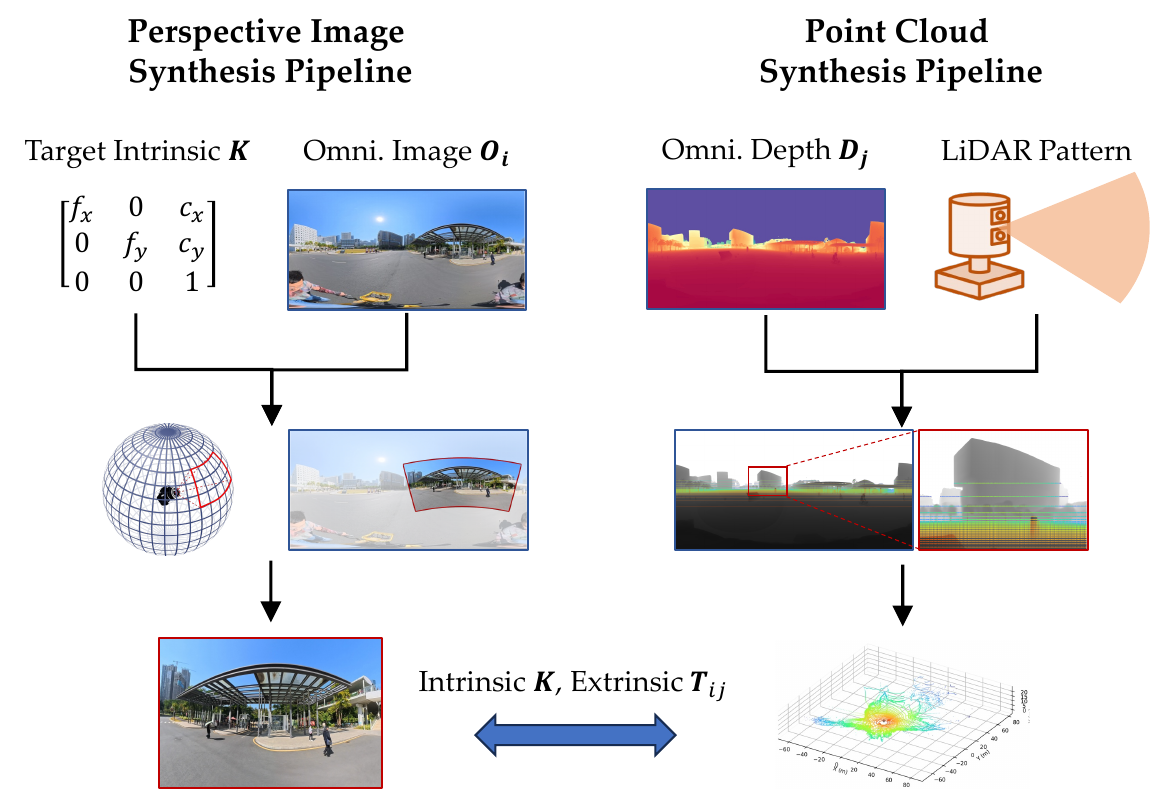}
  \caption{The proposed \textit{Omni2Perspective} and \textit{Omni2LiDAR} data synthesis pipelines.}
  \label{fig_synthesis}
\end{figure}

\subsubsection{Omni2Perspective}
To synthesize perspective images, we first utilize the rotation component of the
camera's extrinsic parameters to determine the coordinates of the perspective image's
origin (principal point) on the ERP image.
Given the intrinsic parameters of the target perspective camera, including the focal
length $\{f_x, f_y\}$ and image dimensions $W_{\text{crop}} \times H_{\text{crop}}$,
we calculate the equivalent horizontal and vertical FoVs.
Since the pixel resolution of the ERP image directly maps to the angular resolution,
we can determine the exact sampling dimensions required for these intrinsics.
After extracting this region, the image is resized to the target output shape of
$W_{\text{out}} \times H_{\text{out}}$.

\subsection{Extrinsic-constrained Data Pair Generation}
\label{subsect_par_generation}
Given panoramic images $\mathbf{O}_i$ and $\mathbf{O}_j$ with a known relative pose 
$\mathbf{T}_{O_i}^{O_j} = \{\mathbf{R}_{O_i}^{O_j}, \mathbf{t}_{O_i}^{O_j}\}$, 
we can synthesize virtual LiDAR $\mathbf{L}$ and camera $\mathbf{C}$ data at their 
respective locations. 
To ensure the generated pairs strictly satisfy the target extrinsics 
$\mathbf{T}_L^C = \{\mathbf{R}_L^C, \mathbf{t}_L^C\}$, geometric constraints must 
be established. 
The standard rigid transformation on a 3D point $\mathbf{p}$ follows a \textit{rotate-then-translate} 
chain as $\mathbf{p}^C = \mathbf{R}_L^C \mathbf{p}^L + \mathbf{t}_L^C$.
Since our virtual sensors share the optical centers of their source panoramas, 
we alter this transformation via a \textit{rotate-translate-rotate} kinematic chain:
\begin{align}
  \mathbf{p}^C &= \mathbf{R}_{O_j}^C \left( \mathbf{R}_{O_i}^{O_j} (\mathbf{R}_L^{O_i} \mathbf{p}^L) + \mathbf{t}_{O_i}^{O_j} \right)  \nonumber \\
               &= \underbrace{(\mathbf{R}_{O_j}^C \mathbf{R}_{O_i}^{O_j} \mathbf{R}_L^{O_i})}_{\mathbf{R}_L^C} \mathbf{p}^L + \underbrace{\mathbf{R}_{O_j}^C \mathbf{t}_{O_i}^{O_j}}_{\mathbf{t}_L^C}.
  \label{eq_kinematic_chain_full}
\end{align}

\begin{figure}
  \centering
  \includegraphics[width=0.98\columnwidth]{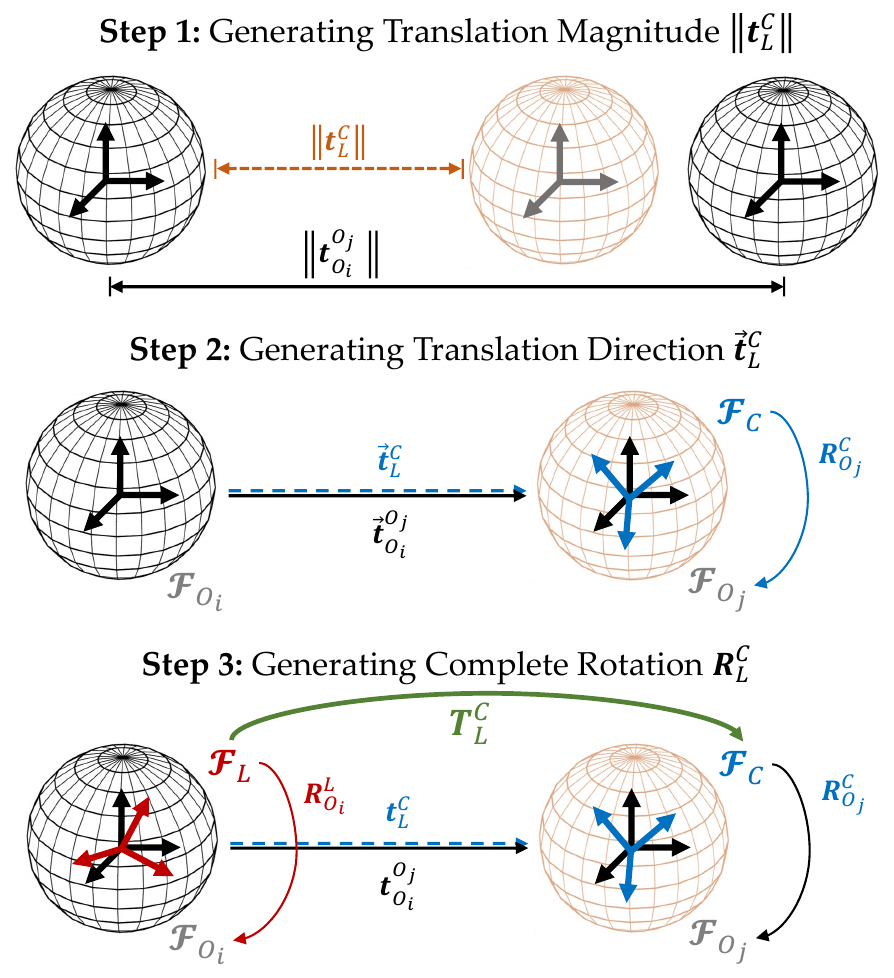}
  \caption{The proposed $3$-step extrinsic-constrained data generation pipeline.}
  \label{fig_3step}
\end{figure}

\subsubsection{3-step Generation Pipeline}
\label{subsect_three_step}
We propose a three-step solution strategy to introduce the extrinsic constraints
in the generation process.

\textbf{Step 1 -- Generating Translation Magnitude $||\mathbf{t}_L^C||$:}
Since the physical baseline of $\mathbf{O}_i$ and $\mathbf{O}_j$ in the dataset is fixed,
we introduce a scale factor $s$ to match the magnitude of the target translation.
The scale factor is defined as:
\begin{equation}
  s = \|\mathbf{t}_L^C\| / \|\mathbf{t}_{O_i}^{O_j}\|.
\end{equation}
According to the principles of visual triangulation,
scaling the entire 3D scene by changing the depth map $\mathbf{D}_i$ to $\mathbf{D}_i^{(s)}=s\mathbf{D}_i$
proportionally rescales the translation baseline to $||\mathbf{t}_L^C||$ without
affecting any rotational components.

\textbf{Step 2 -- Generating Translation Direction $\vec{\mathbf{t}}_L^C$:}
With the result of Eq.~\ref{eq_kinematic_chain_full}, we can isolate the translation 
component as $\mathbf{t}_L^C = \mathbf{R}_{O_j}^C \mathbf{t}_{O_i}^{O_j}$. 
This reveals that the target translation direction $\vec{\mathbf{t}}_L^C = \mathbf{t}_L^C/\|\mathbf{t}_L^C\|$ 
is solely determined by rotating the direction of omnidirectional baseline 
$\vec{\mathbf{t}}_{O_i}^{O_j}=\mathbf{t}_{O_i}^{O_j}/\|\mathbf{t}_{O_i}^{O_j}\|$ 
using the virtual camera's orientation $\mathbf{R}_{O_j}^C$:
\begin{equation}
  \vec{\mathbf{t}}_L^C = \mathbf{R}_{O_j}^C \vec{\mathbf{t}}_{O_i}^{O_j}.
  \label{eq_translation_direction}
\end{equation}

\textbf{Step 3 -- Generating Complete Rotation $\mathbf{R}_L^C$:}
Given $\mathbf{R}_{O_j}^C$ obtained from Step 2 and the known baseline rotation
$\mathbf{R}_{O_i}^{O_j}$, we can analytically derive the required LiDAR orientation
$\mathbf{R}_L^{O_i}$ to satisfy the target extrinsic constraints.
The relationship can be expressed as:
\begin{equation}
  \mathbf{R}_L^{O_i} = (\mathbf{R}_{O_i}^{O_j})^{-1} (\mathbf{R}_{O_j}^C)^{-1} \mathbf{R}_L^C.
  \label{eq_kinematic_chain}
\end{equation}

\subsubsection{Rotational Data Augmentation}
\label{subsubsect_data_augmentation}
While Eq.~(\ref{eq_translation_direction}) establishes the directional constraint,
aligning two 3D unit vectors only consumes $2$ degrees of freedom (DoF).
Since the rotation matrix $\mathbf{R}_{O_j}^C \in \mathrm{SO(3)}$ possesses $3$ DoF,
the kinematic null space of this constraint is $1$-dimensional.

To rigorously derive this, let $\mathbf{R}_{\text{base}}$ be a particular shortest-path
rotation solution that satisfies $\vec{\mathbf{t}}_L^C = \mathbf{R}_{\text{base}} \vec{\mathbf{t}}_{O_i}^{O_j}$.
We introduce a rotational perturbation in the Lie group, represented by $\exp([\boldsymbol{\omega}]_{\times})$,
where $\boldsymbol{\omega} \in \mathfrak{so}(3)$ is the Lie algebra of $\mathrm{SO(3)}$
and $[\cdot]_{\times}$ denotes the skew-symmetric matrix operator.
The perturbed rotation $\mathbf{R}_{O_j}^C = \exp([\boldsymbol{\omega}]_{\times}) \mathbf{R}_{\text{base}}$
must still satisfy Eq.~(\ref{eq_translation_direction}) and we can also introduce it into Eq.~(\ref{eq_translation_direction}) as
\begin{equation}
  \exp([\boldsymbol{\omega}]_{\times}) \mathbf{R}_{\text{base}} \vec{\mathbf{t}}_{O_i}^{O_j} = \vec{\mathbf{t}}_L^C \implies \exp([\boldsymbol{\omega}]_{\times}) \vec{\mathbf{t}}_L^C = \vec{\mathbf{t}}_L^C.
\end{equation}
Applying the first-order Taylor expansion for a small perturbation, $\exp([\boldsymbol{\omega}]_{\times}) \approx \mathbf{I} + [\boldsymbol{\omega}]_{\times}$,
the condition simplifies to:
\begin{align}
  (\mathbf{I} + [\boldsymbol{\omega}]_{\times}) \vec{\mathbf{t}}_L^C & = \vec{\mathbf{t}}_L^C, \nonumber \\
  \implies [\boldsymbol{\omega}]_{\times} \vec{\mathbf{t}}_L^C       & = \mathbf{0}, \nonumber           \\
  \implies \boldsymbol{\omega} \times \vec{\mathbf{t}}_L^C           & = \mathbf{0}.
\end{align}
This indicates that $\boldsymbol{\omega}$ must be parallel to $\vec{\mathbf{t}}_L^C$.
An intuitive explanation can be derived from Eq.~\ref{eq_kinematic_chain_full}:
In its rotational part, various combinations of $\mathbf{R}_{O_j}^C$ and $\mathbf{R}_{L}^{O_i}$ 
can satisfy a fixed rotation $\mathbf{R}_{L}^C$.
In the translation part, an additional rotation superimposed on the direction of 
$\mathbf{t}_{O_i}^{O_j}$ can also guarantee the invariance of $\mathbf{t}_{L}^{C}$.
The null space can be parameterized by a single scalar angle $\phi$,
such that $\boldsymbol{\omega} = \phi \vec{\mathbf{t}}_L^C$.
The general solution for the camera orientation is thus given by:
\begin{equation}
  \mathbf{R}_{O_j}^C = \mathbf{R}_{\text{null}}(\phi, \vec{\mathbf{t}}_L^C) \mathbf{R}_{\text{base}},
  \label{eq_general_solution}
\end{equation}
where $\mathbf{R}_{\text{null}}(\phi, \vec{\mathbf{t}}_L^C)$ represents a rotation
by an arbitrary angle $\phi \in [0, 2\pi)$ around the axis $\vec{\mathbf{t}}_L^C$.

We leverage this $1$-DoF null space as a mathematically rigorous data augmentation strategy.
During the data generation process, we uniformly sample the angle $\phi \sim \mathcal{U}(0, 2\pi)$
to compute $\mathbf{R}_{O_j}^C$, which in turn yields diverse valid solutions for
$\mathbf{R}_L^{O_i}$ via Eq.~(\ref{eq_kinematic_chain}).
This allows us to synthesize an infinite number of valid LiDAR-camera pairs with extensive
viewpoint rotations while strictly preserving the required extrinsic constraints.

\subsection{Data Pair Selection and Post Filtering}
\label{subsect_post_process}
To enhance the efficiency of data generation and ensure the high quality of the
synthesized dataset, we implement a rigorous two-stage filtering mechanism:
a pre-synthesis candidate selection on the omnidirectional data, and a post-synthesis
verification on the generated pairs.

Before synthesis (Section~\ref{subsect_par_generation}), we coarsely filter omnidirectional
pairs ($\mathbf{O}_1$ and $\mathbf{O}_2$), retaining those with a relative translation
below $\tau_{\text{dist}} = 5$~m.
Next, a fine-grained co-visibility check is performed via bidirectional projection
using the relative pose $\mathbf{T}_{O_1}^{O_2}$ and depth maps ($\mathbf{D}_1$ and $\mathbf{D}_2$).
When projecting $\mathbf{O}_1$ onto $\mathbf{O}_2$, a Z-buffer strategy~\cite{cattaneo2025tro}
resolves occlusions by retaining only the minimum depth.
The projected depth $\mathbf{D}_1^{\text{proj}}$ is then pixelwisely compared against
$\mathbf{D}_2$.
Pixels with depth ratio discrepancies $\tau_{\text{diff}} < 0.05$ are deemed
valid.
To guarantee sufficient cross-modal information overlap, the valid overlapping
ratio $\gamma_{\text{valid}}$ is quantitatively defined as:
\begin{equation}
  \gamma_{\text{valid}} = \frac{1}{N} \sum_{p \in \Omega} \mathbb{I} \left( \frac{\left| \mathbf{D}_1^{\text{proj}}(p) - \mathbf{D}_2(p) \right|}{\max \left(\mathbf{D}_1^{\text{proj}}(p), \mathbf{D}_2(p) \right) } < \tau_{\text{diff}} \right).
\end{equation}
Data pairs are valid when the pixel ratio $\gamma_{\text{valid}} > 30\%$ in both
projection directions.
Finally, to augment dataset generalization, we randomly assign the virtual LiDAR
and camera roles between $\mathbf{O}_1$ and $\mathbf{O}_2$ during generation.

After generating the virtual LiDAR-camera pairs, the perspective images undergo
significant FOV cropping.
To guarantee data validity within the cropped regions, we apply a post-processing
step involving a unidirectional projection from the virtual LiDAR to the virtual camera.
We re-evaluate the overlap ratio using the aforementioned depth consistency check,
but impose a stricter threshold of $\gamma_{\text{valid}} \ge 80\%$.
Furthermore, recognizing that structural edges provide the most critical geometric cues
for LiDAR-camera registration tasks, we assess the structural richness of the generated pairs.
We apply the Sobel operator to the projected depth map $\mathbf{D}$ to compute the
depth gradient magnitude.
Pixels with a gradient magnitude exceeding $0.03$~m are classified as edge pixels.
A synthesized data pair is preserved only if edge pixels constitute more than $5\%$ of
the entire image, ensuring abundant geometric features for downstream tasks.

\section{Experiment}
\subsection{Datasets and devices}
To generate the training data, we utilize the omnidirectional subset from 
TartanGround~\cite{patel2025iros} dataset, which comprises $63$ diverse scenes.
Instead of using the entire dataset, we deliberately exclude unstructured natural
scenes (e.g., forests or heavily vegetated areas) that contain severe structural
noise and lack distinct geometric features.
We carefully select $20$ high-quality structured scenes with $18861$ images, 
allocating $16$ scenes for training and $4$ scenes for testing.
Following our synthesis pipeline, we generate $2000$ valid LiDAR-camera data pairs per scene.
This yields a comprehensive dataset comprising $32000$ training pairs and $8000$ testing pairs,
of which each pair owns a unique extrinsic parameter.

\begin{figure}
  \centering
  \includegraphics[width=0.95\columnwidth]{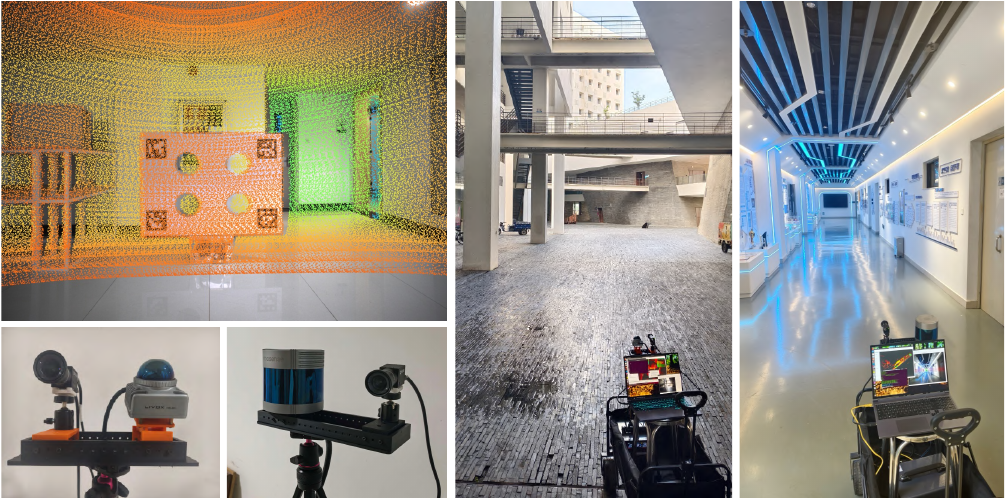}
  \caption{Our real-world rig configurations and experiment scenarios.
  The ground truth extrinsics are derived by using the target-based calibration 
  method of~\cite{zheng2026rap}.
  }
  \label{fig_device}
\end{figure}

To evaluate the capability of our trained model in real-world scenarios,
we construct a custom data collection platform featuring two distinct sensor 
configurations (Fig.~\ref{fig_device}).
Both configurations share the same camera module: a FLIR BFS-U3-31S4C camera equipped
with a lens providing a horizontal FoV of approximately $90^\circ$.
The captured images are resized and cropped to a resolution of $576 \times 768$ to
match the network's input requirements.
For the 3D modality, we employ two different LiDAR sensors to assess the method's
robustness across diverse scanning patterns: a traditional mechanical spinning LiDAR~(RoboSense RS-LiDAR-32)
and a non-repetitive scanning pattern~(Livox Mid-360).

\begin{figure*}
  \centering
  \includegraphics[width=\linewidth]{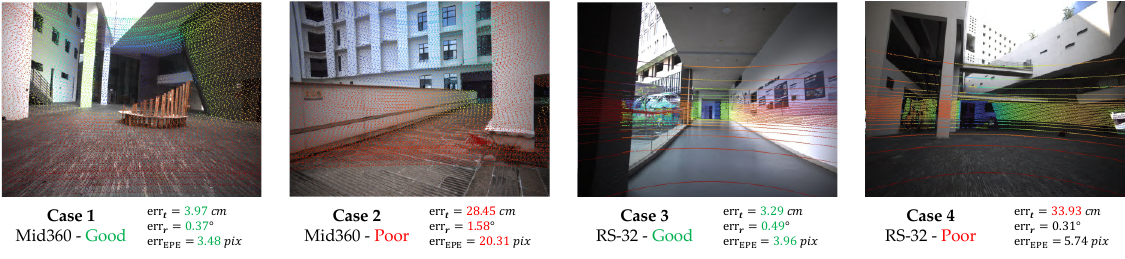}
  \caption{Qualitative analysis of LiDAR-to-camera projections using CMRNext's outputs.
  Online calibration typically achieves high accuracy when the scene is rich in multi-directional 
  edge features (Cases 1 \& 3). 
  However, an uneven spatial distribution of these features can induce slight rotation-translation 
  coupling, leading to increased calibration errors, even if the edges are visually aligned (Case 2). 
  When employing sparse, repetitive scanning LiDAR sensors, the lack of cross-modal overlap 
  exacerbates information loss within the point cloud, ultimately resulting in 
  substantial independent translation errors (Case 4).}
  \label{fig_exp}
\end{figure*}

\subsection{Baseline Methods and Evaluation Criterion}
\label{subsect_baselines_and_metrics}
Since the core contribution of our work focuses on enhancing the generalization
of training data and mitigating learning shortcuts, we benchmark our approach
against the prevailing data generation paradigm.
The standard paradigm typically relies on a single (or a strictly limited set) of
extrinsic calibrations to generate training data, heavily depending on the injection
of random extrinsic perturbations on-the-fly as a form of data augmentation.
In contrast, our proposed method achieves inherent dataset-level generalization
through diverse viewpoint synthesis, while still incorporating minor synthetic
perturbations during training to further augment the data and ensure robustness.

As our work focuses on the data synthesis pipeline rather than architectural innovations,
we validate the efficacy of our dataset using two widely adopted LiDAR-camera calibration
networks representing different paradigms:
\begin{itemize}
  \item \textbf{CMRNext}~\cite{cattaneo2025tro} is a representative geometry-based
        framework. It employs a RAFT-inspired optical flow architecture for dense multimodal
        feature alignment and utilizes a PnP solver for the final geometric pose estimation.
  \item \textbf{CalibNet}~\cite{iyer2018iros} is a classic regression-based network.
        It utilizes a dual-branch encoder to independently extract and compress point cloud
        and image features, culminating in fully connected layers that directly regress
        the 6-DoF extrinsic pose parameters.
\end{itemize}

To quantitatively assess the calibration performance, we employ two primary evaluation
metrics. We first compute the absolute pose error~(APE) between the network's predicted
6-DoF pose $\mathbf{\bar{T}}$ and the ground truth extrinsic $\mathbf{T}_{\text{gt}}$, 
evaluating the translation and rotation deviations, respectively:
\begin{align}
  \text{err}_{R} & = \arccos \left( \frac{\text{Tr}(\mathbf{R}_{\text{gt}}^\top \mathbf{\bar{R}}) - 1}{2} \right),
  \\
  \text{err}_{t} & = \| \mathbf{\bar{t}} - \mathbf{t}_{\text{gt}} \|_2.
\end{align}
For CMRNext that generates intermediate dense alignment representations, we evaluate
the multi-modal alignment quality using the end-point error~(EPE), defined as the
2D pixel discrepancy between the LiDAR points projected using the estimated extrinsics
and those projected using the ground truth:
\begin{equation}
  \text{err}_{\text{EPE}} = \| \pi(\mathbf{\bar{T}} \mathbf{P}_i) - \pi(\mathbf{T}_{\text{gt}} \mathbf{P}_i) \|_2.
\end{equation}
For all metrics, we report both the \textit{mean} and \textit{median} errors across
the test set to provide a comprehensive and robust statistical evaluation.

\subsection{Experiment 1: Synthetic Data Training and Inference}
\label{subsect_synthetic_eval}
We first conduct training and evaluation on our synthesized datasets to establish
a controlled baseline.
To simulate initial misalignments, we inject random extrinsic perturbations uniformly
sampled within a range of $20$~cm per axis for translation and $10^\circ$ per angle for rotation.
During the training phase, we train the CMRNext model for $300$ epochs and the CalibNet
for $30$ epochs, which aligns with their respective convergence characteristics.

\begin{table}
  \centering
  \caption{Cross-evaluation on different data generation paradigms with different LiDAR patterns.
    Values indicate the Mean/Median errors.}
  \label{tab_exp1}
  \resizebox{\columnwidth}{!}{%
    \begin{tabular}{ccccccc}
      \toprule
      LiDAR & Model                   & Train & Test & Rot. Err. (\(^\circ\)) & Trans. Err. (cm) & EPE. Err. (pixel) \\
      \midrule

      \multirow{8}{*}{\rotatebox{90}{Mid-360}}
                 & \multirow{4}{*}{\rotatebox{90}{CMRNext}}
                 & Trad.                     & Trad.  & 0.35 / 0.17    & 1.49 / 0.83   & 4.99 / 2.18     \\
                 &
                 & Trad.                     & Ours   & 4.25 / 3.17    & 18.77 / 16.75 & 61.91 / 46.38   \\
                 &
                 & Ours                      & Ours   & 0.72 / 0.26    & 3.10 / 1.49   & 15.66 / 6.51    \\
                 &
                 & Ours                      & Trad.  & 1.31 / 0.46    & 6.44 / 3.09   & 22.46 / 7.03    \\
      \cmidrule{2-7}

                 & \multirow{4}{*}{\rotatebox{90}{CalibNet}}
                 & Trad.                     & Trad.  & 4.97 / 4.4     & 16.49 / 15.34 & 57.15 / 48.40   \\
                 &
                 & Trad.                     & Ours   & 10.04 / 9.84   & 22.75 / 21.94 & 107.34 / 98.95  \\
                 &
                 & Ours                      & Ours   & 8.35 / 8.23    & 16.97 / 16.85 & 82.51 / 78.87   \\
                 &
                 & Ours                      & Trad.  & 7.98 / 7.80    & 19.95 / 19.97 & 80.45 / 75.89   \\
      \midrule

      \multirow{8}{*}{\rotatebox{90}{RS-LiDAR-32}}
                 & \multirow{4}{*}{\rotatebox{90}{CMRNext}}
                 & Trad.                     & Trad.  & 0.63 / 0.17    &  2.34 / 1.32   & 6.56 / 1.98    \\
                 &
                 & Trad.                     & Ours   & 6.50 / 5.99    & 21.69 / 19.04  & 53.13 / 35.54  \\
                 &
                 & Ours                      & Ours   & 0.99 / 0.34    & 5.15 / 2.80    & 12.11 / 4.25   \\
                 &
                 & Ours                      & Trad.  & 2.75 / 1.00    & 15.05 / 8.40   & 32.90 / 9.70   \\
      \cmidrule{2-7}

                 & \multirow{4}{*}{\rotatebox{90}{CalibNet}}
                 & Trad.                     & Trad.  & 6.20 / 5.81    & 20.20 / 19.20  & 81.13 / 65.48  \\
                 &
                 & Trad.                     & Ours   & 9.40 / 9.32    & 23.11 / 22.93  & 93.51 / 87.34  \\
                 &
                 & Ours                      & Ours   & 11.27 / 9.02   & 21.17 / 16.00  & 165.87 / 81.62 \\
                 &
                 & Ours                      & Trad.  & 9.48 / 8.59    & 22.58 / 20.73  & 108.96 / 85.86 \\
      \bottomrule
    \end{tabular}
  }
\end{table}

To rigorously assess the generalization capabilities of the models, we design a
cross-dataset evaluation protocol.
Specifically, we train the networks independently on datasets generated by two different
paradigms.
Subsequently, we perform cross-inference testing on the test sets of both distributions.
The quantitative results are summarized in Table~\ref{tab_exp1}.
The results show that models trained under the traditional paradigm exhibit a severe
inductive bias and lack generalization capability (especially the experiment sets of 
\textit{Mid360-CMRNext} and \textit{RS-LiDAR-32-CMRNext}).
When the extrinsic parameters in the test set deviate from the narrow distribution
of the training set (e.g., simulating real-world extrinsic shifts from factory defaults),
the inference performance of the traditionally trained models degrades significantly.
In contrast, models trained in our paradigm demonstrate robust and consistent performance
across both test sets, maintaining reasonable and closely aligned accuracy regardless
of the underlying data generation methods.

Interestingly, during the experiments, we observed that networks trained on the traditional
dataset converge faster and are seemingly \textit{easier} to train compared to those
trained on our dataset.
However, rather than indicating superior learning, this apparent ease of training
is a strong indicator of the \textit{dataset shortcut} phenomenon.
In the traditional paradigm, the network exploits spurious correlations and merely
memorizes the restricted extrinsic distribution (the shortcut) instead of learning
the complex, underlying geometric alignment between the point clouds and images.
While this allows the training loss to drop rapidly, it is ultimately detrimental
to the fundamental task of cross-modal calibration, rendering the model fragile in
real-world applications.
Our data generation method effectively mitigates these shortcuts, forcing the network
to learn generalized alignment features.

\subsection{Experiment 2: Real-world Zero-shot Inference}
\label{subsect_zero_shot_eval}
To evaluate the real-world applicability of our approach, we deploy the models trained
exclusively on the synthetic dataset (as described in Section~\ref{subsect_synthetic_eval})
directly onto real-world sensor data.
For each configuration, the real-world data was recorded under three distinct extrinsic 
calibration states, yielding $40$ synchronized LiDAR-camera pairs per state.
The first state, denoted as $\mathbf{T}_0$, represents a factory calibration extrinsic.
During evaluation on $\mathbf{T}_0$, we simulate traditional calibration initialization
by injecting random synthetic perturbations.
The remaining two states ($\mathbf{T}_1$ and $\mathbf{T}_2$), represent slight physical
misalignments introduced relative to $\mathbf{T}_0$.
These physically altered states are specifically designed to evaluate the network's
direct regression performance in authentic, uncalibrated real-world settings without
relying on artificial noise injection.
This zero-shot inference is conducted without any fine-tuning or prior exposure to
the real-world target domain during the training phase.
The quantitative and qualitative results of this evaluation are shown in Table~\ref{tab_exp2}
and Fig.~\ref{fig_exp}.
The performance metrics clearly demonstrate that our proposed data generation endows
the networks with better zero-shot generalization than traditional training paradigm, 
significantly narrowing the domain gap between synthetic data and real-world sensor distributions.
Honestly, we observe that models trained exclusively on synthetic data exhibit a 
pronounced drift in translation errors during zero-shot transfer to real-world scenes. 
A detailed discussion of this phenomenon will be provided in the Section~\ref{sect_diss}.

\begin{table}
  \centering
  \caption{Cross-evaluation on real-world captured data using CMRNext.
    Values indicate the Mean/Median errors.}
  \label{tab_exp2}
  \resizebox{\columnwidth}{!}{%
  \begin{tabular}{cccccc}
    \toprule
    LiDAR & Extr. & Train Pdgm. & Rot. Err. (\(^\circ\)) & Trans. Err. (cm) & EPE. Err. (pixel) \\
    \midrule
  
    \multirow{6}{*}{\rotatebox{90}{Mid-360}}
               & \multirow{2}{*}{\(\mathbf{T}_{0}\)}
               & Trad. & 0.97 / 0.73 & 19.81 / 16.83 & 15.08 / 13.32 \\
               &
               & Ours  & 0.57 / 0.42 & 11.90 / 9.86 & 9.58 / 7.66 \\
    \cmidrule{2-6}
  
               & \multirow{2}{*}{\(\mathbf{T}_{1}\)}
               & Trad. & 1.08 / 0.88 & 23.70 / 21.47 & 14.41 / 12.95 \\
               &
               & Ours  & 0.50 / 0.39 & 13.23 / 11.97 & 9.33 / 8.29 \\
    \cmidrule{2-6}
  
               & \multirow{2}{*}{\(\mathbf{T}_{2}\)}
               & Trad. & 1.15 / 1.01 & 20.75 / 17.55 & 19.64 / 17.61 \\
               &
               & Ours  & 0.58 / 0.44 & 12.16 / 9.98 & 11.18 / 8.31 \\
    \midrule
  
    \multirow{6}{*}{\rotatebox{90}{RS-LiDAR-32}}
               & \multirow{2}{*}{\(\mathbf{T}_{0}\)}
               & Trad. & 0.38 / 0.29 & 12.28 / 9.43 & 6.41 / 4.28 \\
               &
               & Ours  & 0.70 / 0.48 &  15.79 / 12.22 & 7.24 / 5.87 \\
    \cmidrule{2-6}
  
               & \multirow{2}{*}{\(\mathbf{T}_{1}\)}
               & Trad. & 5.03 / 1.21 & 16.99 / 12.91 & 57.33 / 22.49 \\
               &
               & Ours  & 1.52 / 1.29 & 14.82 / 12.79 & 12.33 / 9.03 \\
    \cmidrule{2-6}
  
               & \multirow{2}{*}{\(\mathbf{T}_{2}\)}
               & Trad. & 3.17 / 1.58 & 20.96 / 13.14 & 32.73 / 12.53 \\
               &
               & Ours  & 1.73 / 1.50 & 14.80 / 13.26 & 14.77 / 11.74 \\
    \bottomrule
  \end{tabular}
  }
\end{table}

\subsection{Experiment 3: Multi-stage Cascaded Inference}
\label{subsect_cascaded_inference}
To achieve high-precision calibration, the original architecture of CMRNext employs
a multi-stage inference strategy, where models are progressively trained on varying
ranges of extrinsic perturbations and are used for inference as a cascading pipeline.
Following this paradigm, we design a corresponding three-stage training pipeline,
yielding three distinct inference networks specialized for coarse-to-fine alignment.
The output of each stage serves as the initial pose estimate for the subsequent stage,
progressively refining the extrinsic parameters.
As demonstrated in Table~\ref{tab_exp3}, this cascaded approach enables our algorithm 
to ultimately achieve a competitive calibration accuracy (a median error 
of $1.06$~cm and $0.24^\circ$, with a mean error of $3.35$~cm and $0.96^\circ$). 

\section{Discussion}
\label{sect_diss}
Our experiments reveal that geometry-based calibration networks (e.g., CMRNext) 
are significantly more robust than direct regression architectures (e.g., CalibNet). 
Historically, regression models severely overfit to specific spatial priors (driving scenes) 
from high-density LiDAR datasets like KITTI. 
Consequently, their performance degrades drastically when LiDAR resolution decreases 
or scene complexity increases (as seen in our \textit{RS-LiDAR-32-CalibNet} subset). 
However, this limitation may not be inherent; scaling up both diverse training data 
and model capacity could potentially enable regression networks to overcome these bottlenecks.

Furthermore, models trained on our synthetic pipeline demonstrate promising zero-shot 
generalization despite the sim-to-real gap. 
This robustness stems from the model's reliance on cross-modal geometric edges 
(low-frequency signals), making it less sensitive to the high-frequency textures 
that typically exacerbate domain shifts. 
Nevertheless, we observe notable translation error perturbations during real-world 
zero-shot deployment. 
This likely results from discrepancies in pairwise image baseline distributions 
between synthetic data and real-world extrinsics. 
Scaling these baselines inadvertently introduces a scene depth bias, compromising 
generalization. 
Future work will leverage advanced simulators and high-fidelity image generation 
to bridge this remaining gap.

\begin{table}
  \centering
  \caption{Performance evaluation at different stages with input perturbations.
  Values indicate the Mean/Median errors. 
  Noted that the input perturbation for translation and rotation is introduced 
  per axis and per angle, respectively.}
  \label{tab_exp3}
  \resizebox{\columnwidth}{!}{
    \begin{tabular}{ccccc}
      \toprule
      Stage & Input Perturb.         & Rot. Err. ($^\circ$) & Trans. Err. (cm) & EPE. Err. (pixel) \\
      \midrule
      1     & [$20$ cm, $10^\circ$]  & 1.35 / 0.51 & 5.03 / 2.71 & 25.94 / 12.52 \\
      2     & [$5$ cm,  $1.5^\circ$] & 1.01 / 0.27 & 3.53 / 1.14 & 15.34 / 3.70 \\
      3     & [$3$ cm,  $0.8^\circ$] & 0.96 / 0.24 & 3.35 / 1.06 & 14.11 / 3.17 \\
      \bottomrule
    \end{tabular}
  }
\end{table}

\section{Conclusion}
In this paper, we addressed the critical \textit{dataset shortcut} phenomenon that 
severely limits the real-world applicability and generalization of existing learning-based 
LiDAR-camera calibration networks. 
To break the networks' reliance on fixed extrinsics, we proposed a highly generalizable 
data generation framework. 
By leveraging omnidirectional images with depth ground truth, our pipeline synthesizes 
highly diverse, extrinsically constrained multimodal training data. 
Extensive experiments confirmed that classical calibration networks trained exclusively 
on our synthesized data exhibit significantly enhanced robustness against extrinsic 
perturbations. 
Our approach demonstrated remarkable zero-shot transfer capabilities across various 
physical sensor configurations, validating its practical viability for real-world 
robotic deployments. 

In future work, we intend to extend our extrinsic data generation pipeline directly to 
real-world captured omnidirectional datasets. 
This advancement aims to fundamentally bridge the sim-to-real domain gap inherent 
in purely synthetic data, while significantly scaling up scene diversity and overall 
data richness. 
Ultimately, by leveraging these massively scaled and highly diverse datasets, we want 
to develop a robust cross-modal alignment foundation model for point clouds and images, 
paving the way for universal, zero-shot sensor calibration across diverse hardware 
configurations.

\section*{Acknowledgment}
During the preparation of this manuscript, the authors used Gemini-3.1-Pro for
text polishing and grammar correction.
After using this tool, the authors thoroughly reviewed and edited the manuscript,
and take full responsibility for the final content of this publication.

\bibliographystyle{IEEEtran}
\bibliography{ref}

@article{lang2026ijrr,
  title   = {Gaussian-LIC2: LiDAR-Inertial-Camera Gaussian Splatting SLAM},
  author  = {Lang, Xiaolei and Lv, Jiajun and Tang, Kai and Li, Laijian and Huang, Jianxin and Liu, Lina and Liu, Yong and Zuo, Xingxing},
  journal = {The International Journal of Robotics Research},
  year    = {2026}
}

@article{li2026ral,
  title   = {LiDAR-VGGT: Cross-Modal Coarse-to-Fine Fusion for Globally Consistent and Metric-Scale Dense Mapping},
  author  = {Wang, Lijie and Guo, Lianjie and Xu, Ziyi and Wang, Qianhao and Gao, Fei and Chen, Xieyuanli},
  journal = {IEEE Robotics and Automation Letters},
  year    = {2026}
}

@inproceedings{liu2022icra,
  title     = {BEVFusion: Multi-Task Multi-Sensor Fusion with Unified Bird's-Eye View Representation},
  author    = {Liu, Zhijian and Tang, Haotian and Amini, Alexander and Yang, Xingyu and Mao, Huizi and Rus, Daniela and Han, Song},
  booktitle = {IEEE International Conference on Robotics and Automation (ICRA)},
  year      = {2023}
}

@article{zheng2024tro,
  author   = {Zheng, Chunran and Xu, Wei and Zou, Zuhao and Hua, Tong and Yuan, Chongjian and He, Dongjiao and Zhou, Bingyang and Liu, Zheng and Lin, Jiarong and Zhu, Fangcheng and Ren, Yunfan and Wang, Rong and Meng, Fanle and Zhang, Fu},
  journal  = {IEEE Transactions on Robotics},
  title    = {FAST-LIVO2: Fast, Direct LiDAR–Inertial–Visual Odometry},
  year     = {2025},
  volume   = {41},
  number   = {},
  pages    = {326-346},
  doi      = {10.1109/TRO.2024.3502198}
}

@inproceedings{koide2023icra,
  title        = {General, single-shot, target-less, and automatic lidar-camera extrinsic calibration toolbox},
  author       = {Koide, Kenji and Oishi, Shuji and Yokozuka, Masashi and Banno, Atsuhiko},
  booktitle    = {2023 IEEE International Conference on Robotics and Automation (ICRA)},
  pages        = {11301--11307},
  year         = {2023},
  organization = {IEEE}
}

@article{geiger2013ijrr,
  title     = {Vision meets robotics: The kitti dataset},
  author    = {Geiger, Andreas and Lenz, Philip and Stiller, Christoph and Urtasun, Raquel},
  journal   = {The international journal of robotics research},
  volume    = {32},
  number    = {11},
  pages     = {1231--1237},
  year      = {2013},
  publisher = {Sage Publications Sage UK: London, England}
}

@inproceedings{caesar2020cvpr,
  title        = {nuscenes: A multimodal dataset for autonomous driving},
  author       = {Caesar, Holger and Bankiti, Varun and Lang, Alex H and Vora, Sourabh and Liong, Venice Erin and Xu, Qiang and Krishnan, Anush and Pan, Yu and Baldan, Giancarlo and Beijbom, Oscar},
  booktitle    = {2020 IEEE/CVF conference on computer vision and pattern recognition (CVPR)},
  pages        = {11618--11628},
  year         = {2020},
  organization = {IEEE}
}

@article{chen2023ral,
  author   = {Chen, Feiyi and Li, Liang and Zhang, Shuyang and Wu, Jin and Wang, Lujia},
  journal  = {IEEE Robotics and Automation Letters},
  title    = {PBACalib: Targetless Extrinsic Calibration for High-Resolution LiDAR-Camera System Based on Plane-Constrained Bundle Adjustment},
  year     = {2023},
  volume   = {8},
  number   = {1},
  pages    = {304-311},
  doi      = {10.1109/LRA.2022.3226026}
}

@article{jiao2023tmech,
  author   = {Jiao, Jianhao and Chen, Feiyi and Wei, Hexiang and Wu, Jin and Liu, Ming},
  journal  = {IEEE/ASME Transactions on Mechatronics},
  title    = {LCE-Calib: Automatic LiDAR-Frame/Event Camera Extrinsic Calibration With a Globally Optimal Solution},
  year     = {2023},
  volume   = {28},
  number   = {5},
  pages    = {2988-2999},
  doi      = {10.1109/TMECH.2023.3259444}
}

@article{zheng2026rap,
  author   = {Zheng, Chunran and Zhang, Fu},
  journal  = {IEEE Robotics and Automation Practice},
  title    = {FAST-Calib: LiDAR-Camera Extrinsic Calibration in One Second},
  year     = {2026},
  volume   = {1},
  number   = {},
  pages    = {108-112},
  doi      = {10.1109/RAP.2026.3692446}
}

@article{huang2024tiv,
  title     = {Online, target-free LiDAR-camera extrinsic calibration via cross-modal mask matching},
  author    = {Huang, Zhiwei and Zhang, Yikang and Chen, Qijun and Fan, Rui},
  journal   = {IEEE Transactions on Intelligent Vehicles},
  volume    = {10},
  number    = {5},
  pages     = {3531--3542},
  year      = {2024},
  publisher = {IEEE}
}

@article{huang2026tro,
  title   = {DST-Calib: A Dual-Path, Self-Supervised, Target-Free LiDAR-Camera Extrinsic Calibration Network},
  author  = {Huang, Zhiwei and Fu, Yanwei and Zhou, Yi and Chen, Xieyuanli and Chen, Qijun and Fan, Rui},
  journal = {arXiv preprint arXiv:2601.01188},
  year    = {2026}
}

@inproceedings{fang2021iros,
  author    = {Fang, Chuan and Ding, Shuai and Dong, Zilong and Li, Honghua and Zhu, Siyu and Tan, Ping},
  booktitle = {2021 IEEE/RSJ International Conference on Intelligent Robots and Systems (IROS)},
  title     = {Single-Shot is Enough: Panoramic Infrastructure Based Calibration of Multiple Cameras and 3D LiDARs},
  year      = {2021},
  volume    = {},
  number    = {},
  pages     = {8890-8897},
  doi       = {10.1109/IROS51168.2021.9636767}
}

@inproceedings{hu2023icra,
  author    = {Hu, Yiyang and Ma, Hui and Jie, Leiping and Zhang, Hui},
  booktitle = {2023 IEEE International Conference on Robotics and Automation (ICRA)},
  title     = {DEdgeNet: Extrinsic Calibration of Camera and LiDAR with Depth-discontinuous Edges},
  year      = {2023},
  volume    = {},
  number    = {},
  pages     = {11439-11445},
  doi       = {10.1109/ICRA48891.2023.10160910}
}

@inproceedings{ye2024iros,
  author    = {Ye, Tianyong and Xu, Wei and Zheng, Chunran and Cui, Yukang},
  booktitle = {2024 IEEE/RSJ International Conference on Intelligent Robots and Systems (IROS)},
  title     = {MFCalib: Single-shot and Automatic Extrinsic Calibration for LiDAR and Camera in Targetless Environments Based on Multi-Feature Edge},
  year      = {2024},
  volume    = {},
  number    = {},
  pages     = {864-871},
  doi       = {10.1109/IROS58592.2024.10802395}
}

@article{xie2022ral,
  author   = {Xie, Yusen and Deng, Lei and Sun, Ting and Fu, Yeyu and Li, Jian and Cui, Xinglong and Yin, Hanxi and Deng, Shuixin and Xiao, Junwei and Chen, Baohua},
  journal  = {IEEE Robotics and Automation Letters},
  title    = {A4LidarTag: Depth-Based Fiducial Marker for Extrinsic Calibration of Solid-State Lidar and Camera},
  year     = {2022},
  volume   = {7},
  number   = {3},
  pages    = {6487-6494},
  doi      = {10.1109/LRA.2022.3173033}
}

@inproceedings{zhu2023icra,
  author    = {Zhu, Jiangtong and Xue, Jianru and Zhang, Pu},
  booktitle = {2023 IEEE International Conference on Robotics and Automation (ICRA)},
  title     = {CalibDepth: Unifying Depth Map Representation for Iterative LiDAR-Camera Online Calibration},
  year      = {2023},
  volume    = {},
  number    = {},
  pages     = {726-733},
  doi       = {10.1109/ICRA48891.2023.10161575}
}

@article{duan2023ral,
  title     = {A robust LiDAR-camera self-calibration via rotation-based alignment and multi-level cost volume},
  author    = {Duan, Zaipeng and Hu, Xuzhong and Ding, Junfeng and An, Pei and Huang, Xiao and Ma, Jie},
  journal   = {IEEE Robotics and Automation Letters},
  volume    = {9},
  number    = {1},
  pages     = {627--634},
  year      = {2023},
  publisher = {IEEE}
}

@article{jung2026ral,
  title   = {{Targetless LiDAR-Camera Calibration with Neural Gaussian Splatting}},
  author  = {Jung, Haebeom and Kim, Namtae and Kim, Jungwoo and Park, Jaesik},
  journal = {IEEE Robotics and Automation Letters},
  volume  = {11},
  number  = {4},
  pages   = {4777--4784},
  year    = {2026},
  doi     = {10.1109/LRA.2026.3665066}
}

@article{ou2023sj,
  author  = {Ou, Ni and Cai, Hanyu and Yang, Jiawen and Wang, Junzheng},
  journal = {IEEE Sensors Journal},
  title   = {Targetless Extrinsic Calibration of Camera and Low-Resolution 3-D LiDAR},
  year    = {2023},
  volume  = {23},
  number  = {10},
  pages   = {10889-10899},
  doi     = {10.1109/JSEN.2023.3263833}
}

@article{ou2026ral,
  author   = {Ou, Ni and Chen, Zhuo and Zhang, Xinru and Wang, Junzheng},
  journal  = {IEEE Robotics and Automation Letters},
  title    = {Native-Domain Cross-Attention for Camera-LiDAR Extrinsic Calibration Under Large Initial Perturbations},
  year     = {2026},
  volume   = {11},
  number   = {6},
  pages    = {6983-6990},
  doi      = {10.1109/LRA.2026.3683597}
}

@article{petek2024ral,
  title     = {Automatic target-less camera-LiDAR calibration from motion and deep point correspondences},
  author    = {Petek, K{\"u}rsat and V{\"o}disch, Niclas and Meyer, Johannes and Cattaneo, Daniele and Valada, Abhinav and Burgard, Wolfram},
  journal   = {IEEE Robotics and Automation Letters},
  volume    = {9},
  number    = {11},
  pages     = {9978--9985},
  year      = {2024},
  publisher = {IEEE}
}

@inproceedings{huang2025acmmm,
  title     = {What really matters for learning-based LiDAR-camera calibration},
  author    = {Huang, Shujuan and Lin, Chunyu and Zhao, Yao},
  booktitle = {Proceedings of the 7th ACM International Conference on Multimedia in Asia},
  pages     = {1--8},
  year      = {2025}
}

@inproceedings{sattler2019cvpr,
  author    = {Sattler, Torsten and Zhou, Qunjie and Pollefeys, Marc and Leal-Taixé, Laura},
  booktitle = {2019 IEEE/CVF Conference on Computer Vision and Pattern Recognition (CVPR)},
  title     = {Understanding the Limitations of CNN-Based Absolute Camera Pose Regression},
  year      = {2019},
  volume    = {},
  number    = {},
  pages     = {3297-3307},
  doi       = {10.1109/CVPR.2019.00342}
}

@article{cattaneo2025tro,
  title   = {CMRNext: Camera to LiDAR Matching in the Wild for Localization and Extrinsic Calibration},
  author  = {Daniele Cattaneo and Abhinav Valada},
  journal = {IEEE Transactions on Robotics},
  year    = {2025},
  volume  = {41},
  pages   = {1995-2013},
  doi     = {10.1109/TRO.2025.3546784}
}

@article{pandey2015jfr,
  title     = {Automatic extrinsic calibration of vision and lidar by maximizing mutual information},
  author    = {Pandey, Gaurav and McBride, James R and Savarese, Silvio and Eustice, Ryan M},
  journal   = {Journal of Field Robotics},
  volume    = {32},
  number    = {5},
  pages     = {696--722},
  year      = {2015},
  publisher = {Wiley Online Library}
}

@inproceedings{stutts2024icra,
  title        = {Mutual information-calibrated conformal feature fusion for uncertainty-aware multimodal 3d object detection at the edge},
  author       = {Stutts, Alex C and Erricolo, Danilo and Ravi, Sathya and Tulabandhula, Theja and Trivedi, Amit Ranjan},
  booktitle    = {2024 IEEE international conference on robotics and automation (ICRA)},
  pages        = {2029--2035},
  year         = {2024},
  organization = {IEEE}
}

@article{wang2026neurips,
  title   = {Moge-2: Accurate monocular geometry with metric scale and sharp details},
  author  = {Wang, Ruicheng and Xu, Sicheng and Dong, Yue and Deng, Yu and Xiang, Jianfeng and Lv, Zelong and Sun, Guangzhong and Tong, Xin and Yang, Jiaolong},
  journal = {Advances in Neural Information Processing Systems},
  volume  = {38},
  pages   = {35928--35959},
  year    = {2026}
}

@article{xue2026tase,
  author   = {Xue, Dabin and Li, Baoquan and Li, Yueyuan and Sun, Fuyun and Zhang, Xuebo},
  journal  = {IEEE Transactions on Automation Science and Engineering},
  title    = {LWFusionFormer: Lightweight Convolution and Transformer Fusion for Accurate and Efficient Depth Estimation},
  year     = {2026},
  volume   = {23},
  number   = {},
  pages    = {9376-9387},
  doi      = {10.1109/TASE.2026.3689768}
}

@article{wang2025taes,
  author   = {Wang, Youwei and Tong, Yunze and Wang, Runhua and Zhang, Shiyong and Song, Zhixing and Zhang, Xuebo},
  journal  = {IEEE Transactions on Aerospace and Electronic Systems},
  title    = {SPTL-LCC: Single-Shot, Pixel-Level, Target-Free, and LiDAR-Type Agnostic LiDAR-Camera Extrinsic Self-Calibration},
  year     = {2025},
  volume   = {61},
  number   = {6},
  pages    = {17567-17583},
  doi      = {10.1109/TAES.2025.3604408}
}

@inproceedings{huang2022360vo,
  title        = {360vo: Visual odometry using a single 360 camera},
  author       = {Huang, Huajian and Yeung, Sai-Kit},
  booktitle    = {2022 international conference on robotics and automation (icra)},
  pages        = {5594--5600},
  year         = {2022},
  organization = {IEEE}
}

@inproceedings{patel2025iros,
  title        = {Tartanground: A large-scale dataset for ground robot perception and navigation},
  author       = {Patel, Manthan and Yang, Fan and Qiu, Yuheng and Cadena, Cesar and Scherer, Sebastian and Hutter, Marco and Wang, Wenshan},
  booktitle    = {2025 IEEE/RSJ International Conference on Intelligent Robots and Systems (IROS)},
  pages        = {20524--20531},
  year         = {2025},
  organization = {IEEE}
}

@inproceedings{iyer2018iros,
  title        = {CalibNet: Geometrically supervised extrinsic calibration using 3D spatial transformer networks},
  author       = {Iyer, Ganesh and Ram, R Karnik and Murthy, J Krishna and Krishna, K Madhava},
  booktitle    = {2018 IEEE/RSJ International Conference on Intelligent Robots and Systems (IROS)},
  pages        = {1110--1117},
  year         = {2018},
  organization = {IEEE}
}

\end{document}